\documentclass[graybox]{svmult}

\usepackage{mathptmx}       
\usepackage{helvet}         
\usepackage{courier}        
\usepackage{type1cm}        
\usepackage{makeidx}         
\usepackage{graphicx}        
\usepackage{multicol}        
\usepackage[bottom]{footmisc}
\usepackage{url}
\usepackage{multirow}
\usepackage{bm}

\usepackage{newtxtext,newtxmath}
\usepackage{hyperref}

\makeindex             

\begin{document}

\title*{An Analysis of User Behaviors for Objectively Evaluating Spoken Dialogue Systems}

\author{Koji Inoue, Divesh Lala, Keiko Ochi, Tatsuya Kawahara and Gabriel Skantze}
\institute{Koji Inoue \at Kyoto University, Japan, \email{inoue.koji.3x@kyoto-u.ac.jp}
\and Divesh Lala \at Kyoto University, Japan, \email{lala@sap.ist.i.kyoto-u.ac.jp}
\and Keiko Ochi \at Kyoto University, Japan, \email{ochi.keiko.5f@kyoto-u.ac.jp }
\and Tatsuya Kawahara \at Kyoto University, Japan, \email{kawahara@i.kyoto-u.ac.jp}
\and Gabriel Skantze \at KTH Royal Institute of Technology, Sweden, \email{skantze@kth.se}
\\
\\
This paper has been accepted for presentation at International Workshop on Spoken Dialogue Systems Technology 2024 (IWSDS 2024) and represents the author's version of the work.
}

%
%
\maketitle

\abstract*{
Establishing evaluation schemes for spoken dialogue systems is important, but it can also be challenging.
While subjective evaluations are commonly used in user experiments, objective evaluations are necessary for research comparison and reproducibility.
To address this issue, we propose a framework for indirectly but objectively evaluating systems based on users' behaviors.
In this paper, to this end, we investigate the relationship between user behaviors and subjective evaluation scores in social dialogue tasks: attentive listening, job interview, and first-meeting conversation.
The results reveal that in dialogue tasks where user utterances are primary, such as attentive listening and job interview, indicators like the number of utterances and words play a significant role in evaluation.
Observing disfluency also can indicate the effectiveness of formal tasks, such as job interview.
On the other hand, in dialogue tasks with high interactivity, such as first-meeting conversation, behaviors related to turn-taking, like average switch pause length, become more important.
These findings suggest that selecting appropriate user behaviors can provide valuable insights for objective evaluation in each social dialogue task.
}

\abstract{
Establishing evaluation schemes for spoken dialogue systems is important, but it can also be challenging.
While subjective evaluations are commonly used in user experiments, objective evaluations are necessary for research comparison and reproducibility.
To address this issue, we propose a framework for indirectly but objectively evaluating systems based on users' behaviors.
In this paper, to this end, we investigate the relationship between user behaviors and subjective evaluation scores in social dialogue tasks: attentive listening, job interview, and first-meeting conversation.
The results reveal that in dialogue tasks where user utterances are primary, such as attentive listening and job interview, indicators like the number of utterances and words play a significant role in evaluation.
Observing disfluency also can indicate the effectiveness of formal tasks, such as job interview.
On the other hand, in dialogue tasks with high interactivity, such as first-meeting conversation, behaviors related to turn-taking, like average switch pause length, become more important.
These findings suggest that selecting appropriate user behaviors can provide valuable insights for objective evaluation in each social dialogue task.
}


\section{Introduction}

In the research and development of spoken dialogue systems (SDSs), establishing evaluation methods is a significant challenge~\cite{walker1997paradise,SIM2015305,abd2020technical,deriu2021survey,zhang2021automatic,ni2023survey}.
The performance of dialogue understanding and response generation has seen remarkable progress in recent years, thanks to the development of large language models (LLMs). The advancement of LLM research and development has been supported by commonly used evaluation methods in the field, along with extensive text dialogue datasets.
Both objective evaluation methods (automatic evaluation) and subjective ones (human evaluation) have been used complementarily. Objective evaluation allows for efficient scalability of evaluation data by enabling automated evaluation measures such as BLEU~\cite{papineni2002bleu} and Distinct~\cite{li2016distinct}. Additionally, using the same evaluation data as other studies ensures comparability and research reproducibility.
On the other hand, subjective evaluation enables a more detailed assessment of each generated system response, capturing aspects that cannot be measured objectively. For instance, evaluating the empathy of responses currently relies on subjective evaluation by humans.
Therefore, in SDSs, it is ideal to enhance the efficiency and reproducibility of research in the field by appropriately utilizing both objective and subjective evaluation methods.

For future research and development of speech dialogue systems, it is important to establish objective evaluation criteria.
For typical task-oriented dialogues, such as restaurant searches, where the dialogue goal is clear, objective evaluation criteria like the accuracy of slot-filling tasks and the success rate of the dialogue have been utilized~\cite{henderson2014dstc,budzianowski2018multiwoz}.
However, the dialogue tasks for SDSs do not always have clearly defined goals. With the advent of conversational AI, SDSs are becoming more realistic in their ability to handle everyday social conversations. This includes brief exchanges like reception and information guidance~\cite{swartout2010ada,iio2020human}, as well as more extended conversations like counseling~\cite{devault2014simsensei,laranjo2018conversational,scoglio2019use,rasouli2022potential} and interviews~\cite{han-etal-2013-counseling,johnston-etal-2013-spoken,yu2019open,inoue2020job}.
In these dialogues, while the purpose of the dialogue can be described clearly, the goal itself is not always explicit and gradually becomes clear through the dialogue, fostering mutual understanding and relationships. Consequently, subjective evaluations have been more commonly employed than objective evaluations in the past.

\begin{figure}[t!]
\centering
\includegraphics[width=70mm]{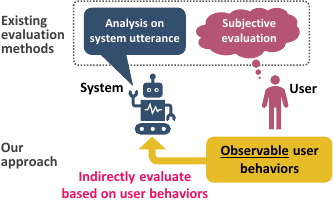}
\caption{Overview of proposed evaluation scheme}
\label{fig:overview}
\end{figure}

In this study, our objective is to establish an objective evaluation method for SDSs in social scenarios where the goals are not clearly defined.
Instead of analyzing system utterances or relying on subjective evaluations from users, we aim to indirectly evaluate the system based on users' spoken-dialogue behaviors during the dialogue, as depicted in Fig.~\ref{fig:overview}.
Users' behaviors in this context, such as the rate of speaking time or the number of spoken words, are objectively observable.
However, it is not clear which behaviors in specific dialogue tasks can be used as clues for evaluation.
To address this, we analyze the relationship between users' behaviors during the dialogue and their subjective evaluations using dialogue data from several social dialogue tasks, including attentive listening, job interview, and first-meeting conversation.
The goal is to identify the behaviors that serve as clues for objective evaluation in different dialogue tasks, which will enable the appropriate selection of users' behaviors for objective evaluation in future research and development.
By measuring and comparing these behaviors, we can achieve the objective evaluation of SDSs across multiple studies, contributing to the overall expansion of the field, which is the ultimate goal of this study.
For example, when comparing two systems, it is ideal to have a situation where not only traditional subjective evaluations are conducted, but also numerical behavior data related to the task is reported.

This paper positions itself as an initial analysis of the relationship between user behavior and subjective evaluation toward the aforementioned goal.
The contributions of this paper are twofold:
\begin{itemize}
    \item Propose an objective evaluation scheme for spoken dialogue systems based on users’ objective behaviors
    \item Clarify the users’ behaviors related to subjective evaluation in social dialogue tasks such as attentive listening, job interview, and first-meeting conversations
\end{itemize}

The rest of this paper is organized as follows.
The proposed evaluation scheme is introduced in Section~\ref{sec:proposed}.
The dialogue data used is explained in Section~\ref{sec:data}.
Then, the relationship between users' behaviors and subjective evaluation is analyzed in Section~\ref{sec:analysis}.
Finally, this paper concludes in Section~\ref{sec:conclusion}.

\section{Proposed Evaluation Scheme} \label{sec:proposed}

The evaluation method proposed in this study is designed to indirectly assess SDSs by analyzing users' behaviors during the dialogue.
The focus is on specific behaviors related to spoken dialogue, including speech, language, and dialogue features.
It is important to note that future research will explore additional modalities, such as eye-gaze behaviors analyzed through image processing.

User behaviors analyzed in this study are listed below.
\begin{itemize}
    \item Utterance time / min.
    \item Number of utterances (IPU segments) / min.
    \item Number of utterance words / min.
    \item Number of unique utterance words / min.
    \item Number of utterance content words / min.
    \item Number of unique utterance content words / min.
    \item Number of backchannels / min.
    \item Number of fillers / min.
    \item Number of laughs / min.
    \item Number of disfluencies / min.
    \item Average switching pause length
\end{itemize}

The criterion for dividing an IPU (inter pausal unit) is set as a silent interval of 200 milliseconds or more.
Since the dialogue data used in this study is in the Japanese language, word segmentation is performed using MeCab\footnote{\url{https://taku910.github.io/mecab}}.
Content words are defined as nouns, verbs, adjectives, adverbs, and conjunctions.
Backchannels are defined as responsive interjections such as ``yes'' or ``uh-huh'', and emotional interjections such as ``hmm'' or ``oh''.
Fillers are expressions used to bridge gaps in conversation, such as ``um'' or ``well'', while speech disfluencies are expressions used for self-correction, such as ``spe, specifically''.
The linguistic behaviors described above were calculated based on manually transcribed data in this study.
Switching pause length refers to the duration of the silent interval when the speaking floor transitions from the system to the user.

Intuitively, the above-mentioned behaviors vary depending on how natural the interaction with the system is.
This can be understood by comparing human-system dialogues and human-human dialogues.
For example, in human-system dialogues, users typically speak clearly and with a limited vocabulary.
On the other hand, in human-human dialogues, it is natural to speak fluently and with a diverse vocabulary.
Regarding backchannels, users in human-system dialogues rarely use them, while in human-human ones, backchannels are used frequently.
Studies have shown that the average switching pause length in human-human dialogues is almost zero seconds~\cite{Levinson2015TurnTaking,skantze2021turn,dingemanse2022text}, while in human-system dialogues, it often takes around 1 to 3 seconds.
Based on these characteristics, evaluating a spoken dialogue system using the above-mentioned behaviors can be seen as evaluating its naturalness and similarity to human behavior.
In this study, we aim to identify specific behaviors related to user subjective evaluation in three different social dialogue tasks.

\section{Dialogue Data} \label{sec:data}

\begin{figure}[t!]
\centering
\includegraphics[width=80mm]{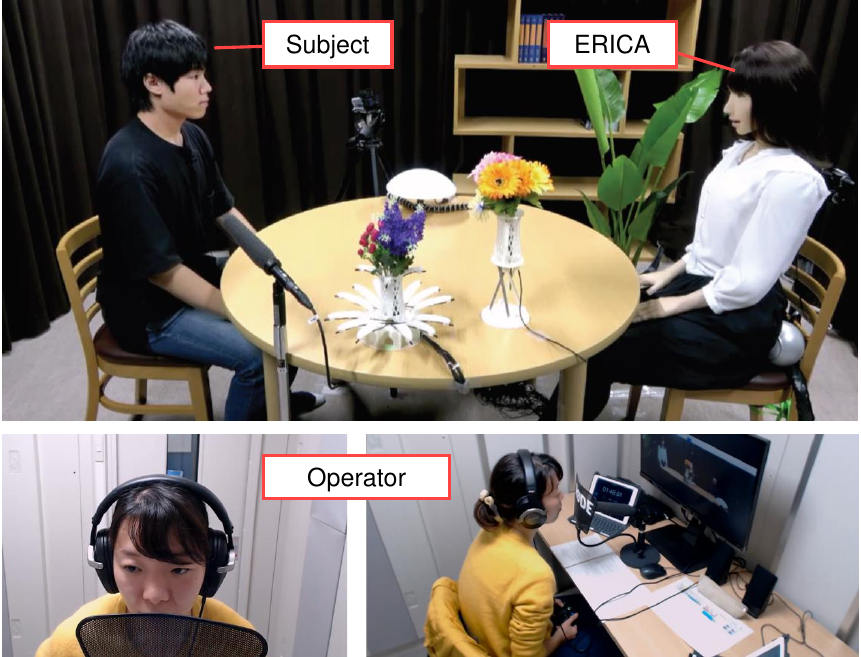}
\caption{Scene of dialogue data collection (Bottom: an operator in WOZ)}
\label{fig:scene}
\end{figure}

The dialogue data used in this study is explained below.
These data were recorded using android ERICA~\cite{inoue-etal-2016-talking}, as shown in Figure \ref{fig:scene}.
It is important to note that in order to introduce variation in the quality of the dialogue, a mixture of dialogues between human-system and human-human (referred to as WOZ: Wizard-of-OZ) was used, depending on the dialogue task.
Table~\ref{table:number} summarizes the number of dialogues used in this study based on the type of setting.
In the case of WOZ, there was an operator in a separate room, and the operator's voice was played through the android's speaker.
Non-verbal expressions, such as the android's gaze and gestures, were controlled by the operator using a handheld controller.
All the dialogues in this study were conducted in Japanese.

In this study, we utilize dialogue data for three different social dialogue tasks, each with its own unique characteristics.
Table~\ref{table:task} provides a summary of these characteristics, specifically in terms of implementing a spoken dialogue system for each task.
Given the variations in the initiative of dialogue, as well as the frequency and clarity of turn-taking, it is anticipated that users' behaviors related to subjective evaluation will also differ.

\begin{table}[t]
  \caption{Number of dialogue in different settings (The names of dialogue tasks are represented as AL, JI, and FMC for Attentive Listening, Job Interview, and First-meeting Conversation, respectively.)}
  \setlength{\tabcolsep}{3mm}
  \label{table:number}
  \begin{center}
  \begin{tabular}{lccc}
  \hline
  \multicolumn{1}{c}{Dialogue task} & AL & JI & FMC \\
  \hline
  \# Autonomous (human-system) & 19 & 86 & 0 \\
  \# WOZ (human-human) & 50 & 0 & 50 \\
  \hline
  Total & 69 & 86 & 50 \\
  \hline
  \end{tabular}
  \end{center}
\end{table}

\begin{table}[t]
  \caption{Characteristics of dialogue tasks targeted in this study}
  \setlength{\tabcolsep}{1.5mm}
  \label{table:task}
  \begin{center}
  \begin{tabular}{lccc}
  \hline
  \multicolumn{1}{c}{Dialogue task} & Attentive Listening & Job Interview & First-meeting Conversation \\
  \hline
  System role & Listening & Asking &  All \\
  Initiative & User & System & Mixed \\
  Majority of utterances & User & User & Both \\
  Majority of backchannels & System & System & Both \\
  Turn-taking & Few & Explicit & Complicated \\
  \hline
  \end{tabular}
  \end{center}
\end{table}

\subsection{Attentive Listening} \label{sec:data:at}

Attentive Listening involves the task of the listener (system) actively listening to the user's talk.
The listener responds through various types of utterances such as backchanneling and elaborating questions.
The authors have developed a real-time spoken dialogue system capable of generating listener responses~\cite{inoue2020sigdial}.
This system was used to record the dialogue data for this study.
We recruited 69 university students as users.
They engaged in an 8-minute dialogue with the attentive listening dialogue system, focusing on the topic of ``difficulties during the COVID-19 pandemic.''
The dialogue task also included human-human dialogue (WOZ).
In total, there were 20 interactions with the autonomous system and 50 WOZ dialogues, resulting in a total of 70 dialogues.

\begin{figure}[t!]
\centering
(a) Attentive Listening\\
\includegraphics[width=90mm]{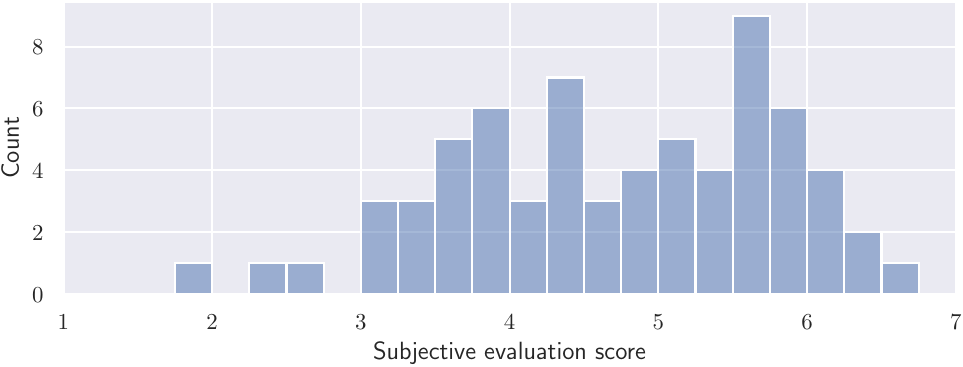}
\vspace{3mm}

(b) Job Interview\\
\includegraphics[width=90mm]{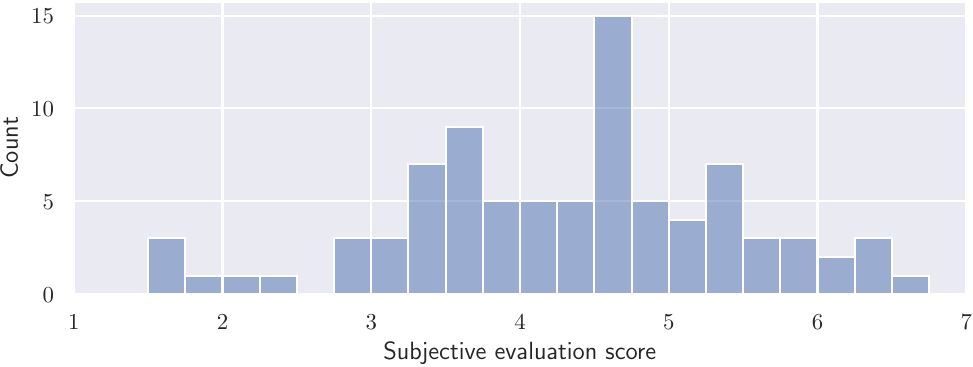}
\vspace{3mm}

(c) First-meeting Conversation\\
\includegraphics[width=90mm]{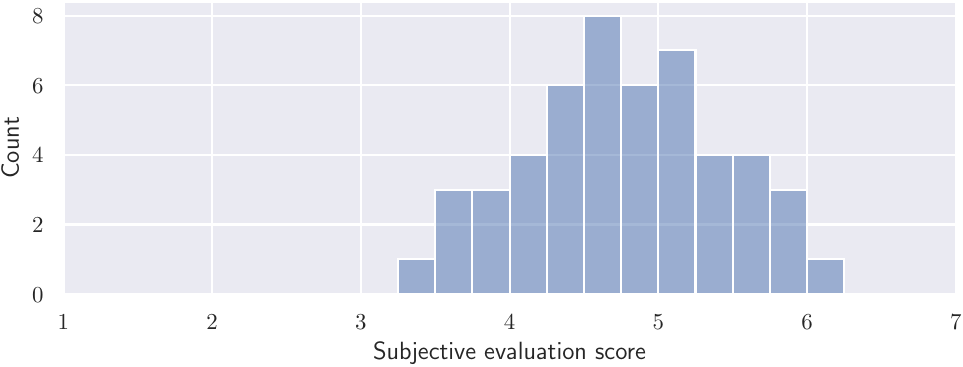}
\caption{Distribution of subjective evaluation scores in each dialogue task}
\label{fig:dist}
\end{figure}

After each dialogue, a subjective evaluation was conducted on the 19 items created in our previous study~\cite{inoue2020sigdial}.
These items include statements such as ``The words uttered by the robot were natural,'' ``The robot understood the talk,'' and ``The robot showed empathy towards me.''
The participants were asked to rate each item on a 7-point scale ranging from 1 to 7.
For this study, we used 18 items, excluding one item that showed variation in interpretation.
We calculated the average value for each dialogue (per participant), which serves as the dependent variable.
In other words, the goal is to predict this average rating value based on the user's behaviors mentioned in the previous section.
Fig.~\ref{fig:dist} (a) shows the distribution of the evaluation scores.
Overall, the scores are somewhat high, but it can be observed that there are also a certain number of participants who gave low scores.

\subsection{Job Interview} \label{sec:data:ji}

Job interview is a dialogue between an interviewer (system) and an applicant (user), where the applicant answers questions posed by the interviewer.
For this study, we utilized a job interview dialogue system developed by the authors~\cite{inoue2020job}, and all interactions were conducted using this autonomous system.
In this task, we treated it as a practice job interview and recruited 43 university students.
Each participant engaged in two times of dialogue with two different systems, which differed in the presence of follow-up questions from the system.
Follow-up questions are inquiries that delve deeper into the applicant's previous response, such as extracting keywords and asking, ``Could you please provide more details about (keyword)?''
Before starting the dialogue, participants were asked to select their desired industry and company, and also prepare answers to several expected questions.

After each dialogue, subjective evaluations were conducted on 19 items created in our previous study~\cite{inoue2020job}.
These items include statements such as ``I was nervous during the interview,'' ``Thanks to the interview, I was able to notice my weak points,'' and ``The interviewer understood my answers.'' Participants were asked to rate each item on a 7-point scale ranging from 1 to 7.
The average value was calculated for each dialogue using the 18 items, excluding the one item that seemed to have variation in interpretation. This average value was then used as the dependent variable.
Fig.~\ref{fig:dist} (b) shows the distribution of the evaluation scores.
It can be seen that there is slightly more variation compared to those of Attentive Listening.

\subsection{First-meeting Conversation} \label{sec:data:ft}

First-meeting conversation is a dialogue that allows both participants to get to know each other and establish a relationship.
For our study, we recruited 50 university and graduate students as users to engage in conversations with a robot, simulating a first-meeting scenario.
Note that the system was completely operated and controlled by the WOZ setup.
Furthermore, the participants were given a list of commonly discussed topics in first-meeting conversations prior to the interaction.

After each dialogue, a subjective evaluation was conducted on 18 items. These items include ``I was able to get to know the other person well,’’ ``The atmosphere of the conversation was pleasant,’’ ``I had a favorable impression of the other person.’’ Participants were asked to rate each item on a 7-point scale, ranging from 1 to 7.
The average value of these 18 items was calculated for each dialogue (participant) and used as the dependent variable.
Fig.~\ref{fig:dist} (c) shows the distribution of the evaluation scores.
It can be observed that the scores are not as distributed as the other two tasks.
The reasons for this could be that all dialogues were performed by the WOZ operator, resulting in overall high quality of the conversations.
Additionally, subjective evaluations of this task are often ambiguous, making it difficult to determine superiority or inferiority.

\section{Analysis} \label{sec:analysis}

The relationship between the subjective evaluation scores mentioned in the previous section and the user behavior mentioned in Section \ref{sec:proposed} was investigated.
To analyze this relationship, we utilized SHAP (SHapley Additive exPlanations)~\cite{lundberg2017shap}\footnote{\url{https://pypi.org/project/shap/}}.
SHAP analysis calculates the contribution level (SHAP value) of each feature in the output value of a trained model.
Specifically, the SHAP value $\phi_f$ of a feature (behavior) $f \in \bm{x}$ is calculated as follows:
\begin{eqnarray}
    \phi_f = \sum_{A \subseteq \bm{x} \setminus \{f\} } \frac{|A|!(|\mathbf{x}|-|A|-1)!}{|\bm{x}|!} ( y(A \cup \{f\}) - y(A) ) \nonumber
\end{eqnarray}
Note that $A$ represents a subset that excludes the feature $f$, and $y(\cdot)$ represents the output of the model when using the given set of features.
In other words, it is the average difference between the output values when using the feature $f$ and when not using it, for all subsets of the features.
The larger the absolute value of this SHAP value, the greater the interpretation that the corresponding feature has a larger impact on the trained model.
One advantage of this analysis method is that it calculates the influence of each feature, taking into account the interaction among the features.
The behaviors described so far are not independent and co-occur during the dialogue, so it is reasonable to analyze them considering the interaction among them, just like SHAP does.

\begin{table}[t!]
\setlength{\tabcolsep}{1mm}
\centering
\begin{tabular}{lcclcclccl}
\hline
\multicolumn{1}{c}{\multirow{2}{*}{User behavior}} & & \multicolumn{8}{c}{Dialogue task} \\
\cline{3-10}
& & \multicolumn{2}{c}{AL} & & \multicolumn{2}{c}{JI} & & \multicolumn{2}{c}{FMC} \\
\hline
Utterance time / min.    && 0.098 & (\phantom{0}7.69) & & \textbf{0.212} & (15.07) & & 0.043 & (\phantom{0}5.42) \\
\# utterances (IPU segments) / min.    && \textbf{0.199} & (15.63) & & \textbf{0.124} & (\phantom{0}8.80) & & \textbf{0.126} & (15.87) \\
\# utterance words / min.   && \textbf{0.162} & (12.72) & & \textbf{0.145} & (10.31) & & 0.070 & (\phantom{0}8.84) \\
\# unique utterance words / min. && \textbf{0.242} & (18.97) & & \textbf{0.183} & (12.98) & & 0.068 & (\phantom{0}8.58) \\
\# utterance content words / min.   && 0.048 & (\phantom{0}3.74) & & 0.087 & (\phantom{0}6.15)  & & 0.016 & (\phantom{0}1.98) \\
\# unique utterance content words / min. && 0.066 & (\phantom{0}5.16) & & 0.042 & (\phantom{0}2.96) & & 0.025 & (\phantom{0}3.21) \\
\# backchannels / min.     && 0.050 & (\phantom{0}3.91) & & 0.067 & (\phantom{0}4.73) & & 0.062 & (\phantom{0}7.83) \\
\# fillers / min.    && \textbf{0.111} & (\phantom{0}8.66) & & \textbf{0.187} & (13.26) & & 0.047 & (\phantom{0}5.99) \\
\# laughs / min.        && 0.093 & (\phantom{0}7.28) & & 0.042 & (\phantom{0}2.98) & & 0.016 & (\phantom{0}2.08) \\
\# disfluencies / min.    && \textbf{0.129} & (10.13) & & \textbf{0.224} & (15.88) & & \textbf{0.192} & (24.23) \\
Average switching pause length    && 0.078 & (\phantom{0}6.10) & & 0.097 & (\phantom{0}6.87) & & \textbf{0.126} & (15.95) \\
\hline 
\end{tabular}
\caption{Absolute mean of SHAP value in each behavior and dialogue task (The numbers in parentheses represent the percentages in each dialogue task. The bold numbers represent the absolute mean SHAP values exceeding 0.100. The names of dialogue tasks are represented as AL, JI, and FMC for Attentive Listening, Job Interview, and First-meeting Conversation, respectively.)}
\label{table:shap}
\end{table}

The procedure for applying SHAP is as follows:
For each dialogue task, we trained a regression model using the user's behavioral features described in Section \ref{sec:proposed} as explanatory variables and the subjective evaluation scores described in Section \ref{sec:data} as the target variable.
XGBoost was used as the regression model in this case.
Then, we computed the SHAP value for each feature using the aforementioned method for the trained model.

\begin{figure*}[t!]
\centering
(a) Attentive Listening
\includegraphics[width=\linewidth]{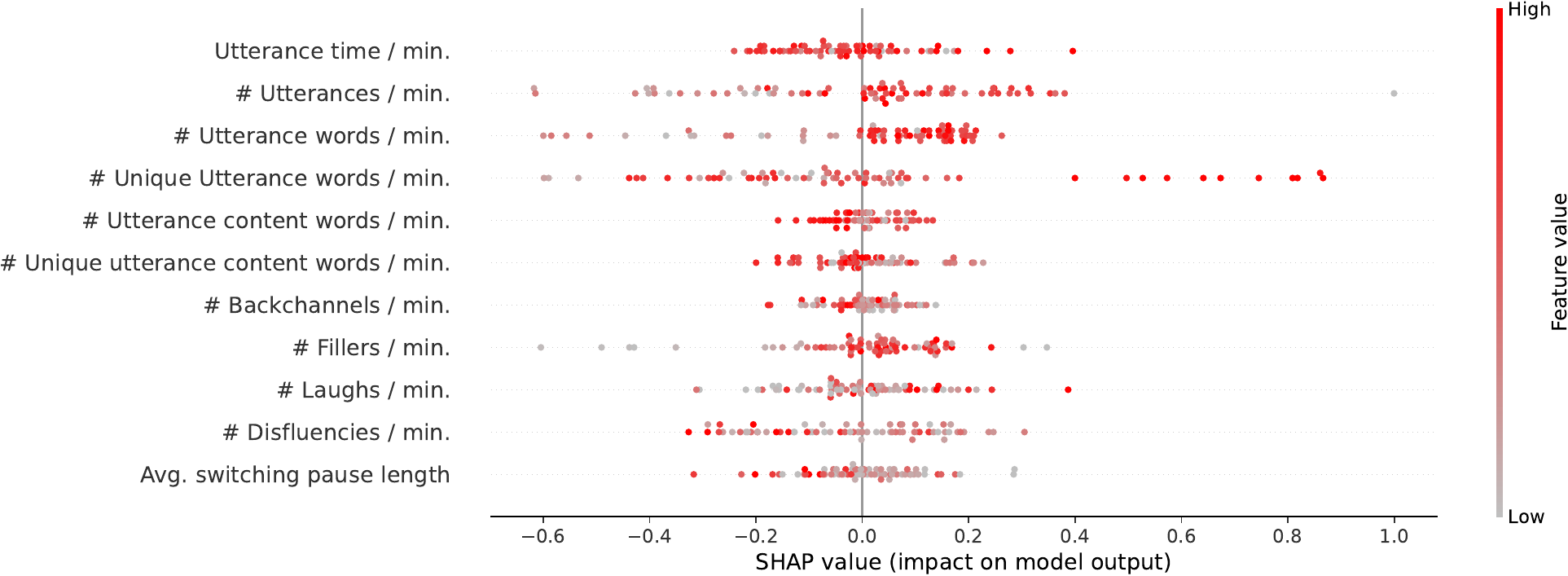}
\vspace{2mm}

(b) Job Interview
\includegraphics[width=\linewidth]{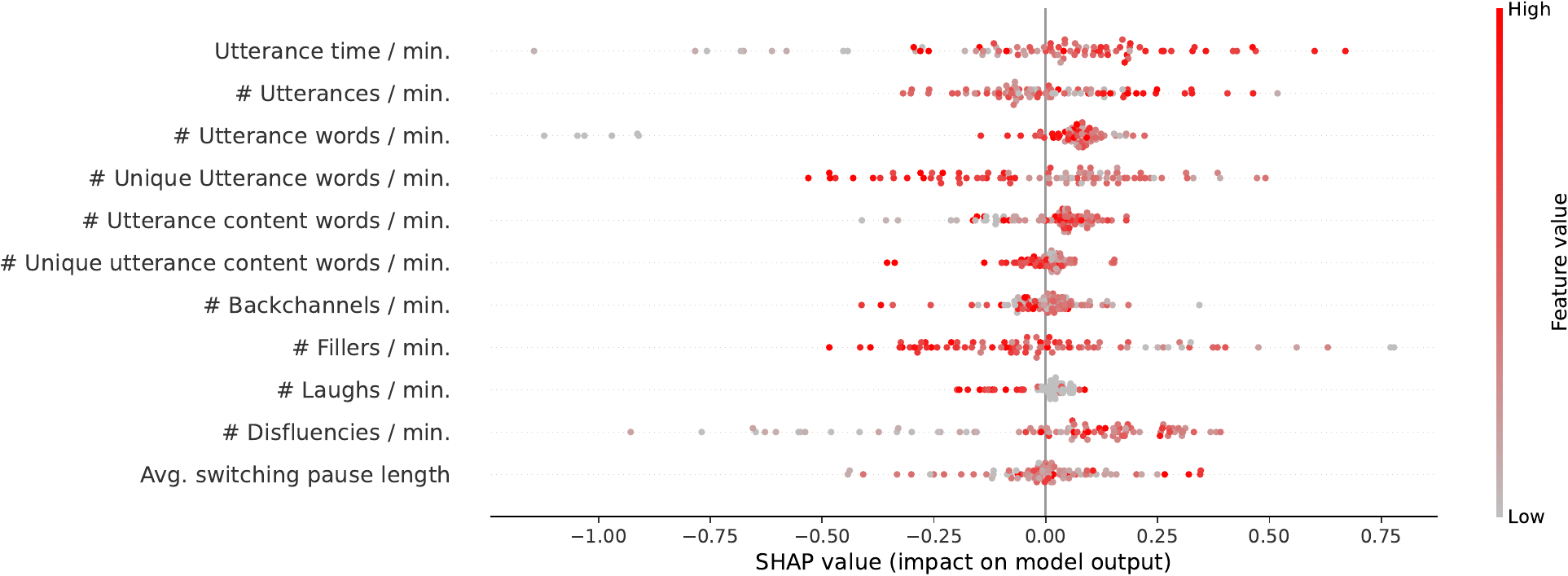}
\vspace{2mm}

(c) First-meeting Conversation
\includegraphics[width=\linewidth]{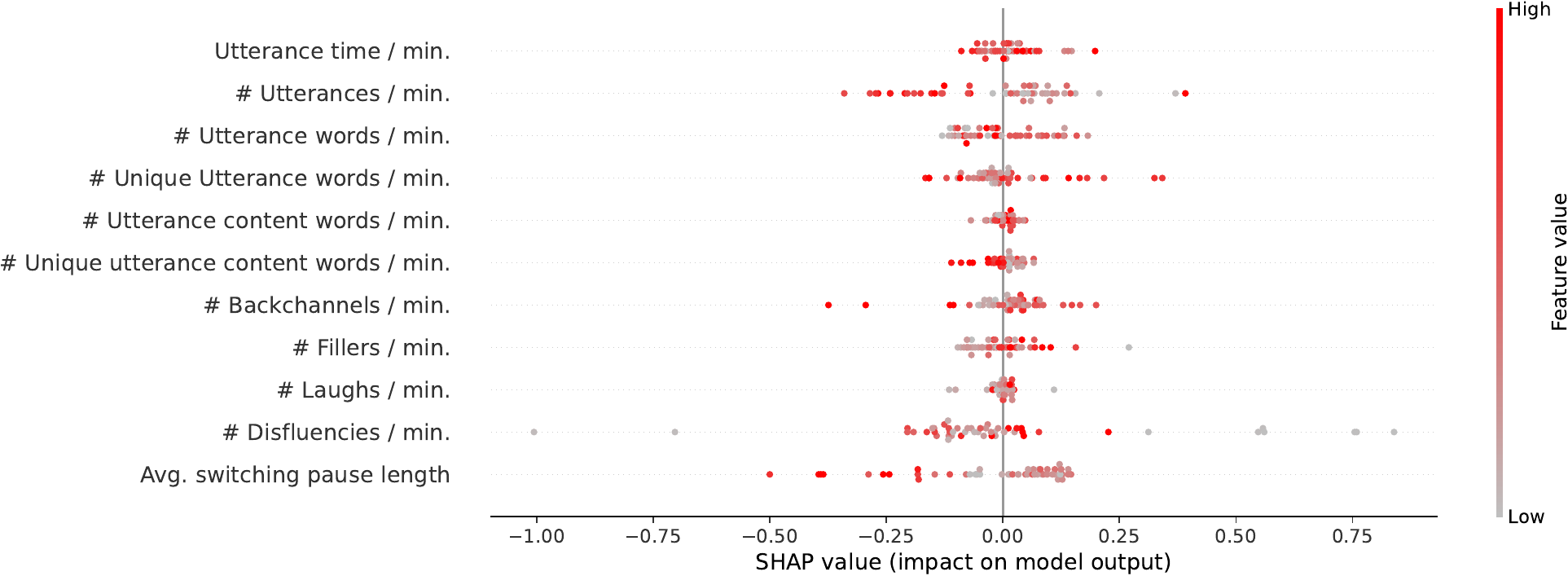}
\caption{Distributions of SHAP value for each behavior and dialogue task}
\label{fig:shap:value}
\end{figure*}

Table~\ref{table:shap} presents the average absolute SHAP value for each behavior in the dialogue tasks.
Across all tasks, the number of utterances (IPU segments) and the number of disfluencies remained consistently high.
When comparing Attentive Listening and Job Interview, similar trends were observed.
Both tasks emphasize the user's speaking ability, leading to the importance of metrics like the number of utterances, words, and those unique numbers.
However, the number of disfluencies was given greater significance in the job interview task.
This is because effective questioning is crucial in Job Interview, so the behavior of disfluencies reflects the evaluation of the system.
For First-meeting Conversation, in addition to the number of utterances and disfluencies, a higher SHAP value was observed for the average switch pause length.
This type of dialogue involves mixed initiative and frequent exchanges, making smooth turn-taking vital for system evaluation.
It should also be noted that First-meeting Conversation, being conducted in the WOZ setting, tends to be more dynamic compared to the autonomous setting so that these interactive factors would be more indicative.
In summary, in dialogue tasks where user utterances are primary, such as Attentive Listening and Job Interview, indicators like the number of utterances and words play a significant role in evaluation.
Additionally, in formal situations like Job Interview, disfluencies are also effective.
On the other hand, in dialogue tasks with high interactivity, such as First-meeting Conversation, the importance of behaviors related to turn-taking, like average switch pause length, increases.

To analyze the specific tendencies of SHAP values, we present the distribution of SHAP values per behavior in each dialogue task in Fig.~\ref{fig:shap:value}. Each point represents a dialogue, and the shading indicates the magnitude of the behavior feature value.
The correlation between the number of utterances in Attentive Listening (Fig.~\ref{fig:shap:value} (a)) and the number of disfluencies in Job Interview (Fig.~\ref{fig:shap:value} (b)) with the magnitude of SHAP values suggests a positive correlation between subjective evaluation scores and these behaviors.
In contrast, the number of unique utterance words in Job Interview (Fig.~\ref{fig:shap:value} (b)), and the number of disfluencies and the average switching pause length in First-meeting Conversation (Fig.~\ref{fig:shap:value} (c)) exhibit an inverse relationship with SHAP values.
Therefore, it can be concluded that there is a negative correlation between subjective evaluation scores and these behaviors.

Finally, we investigated the predictive performance of the regression model trained using leave-one-out cross-validation on subjective evaluation scores.
The input consisted of all the behavioral features (11 dimensions) mentioned earlier, and the output was the subjective evaluation score of the system shown in Figure~\ref{fig:dist}.
XGBoost was used for the model.
When calculating the mean absolute error for the test data, we obtained the following results: 0.970 for Attentive Listening, 0.953 for Job Interview, and 0.683 for First-meeting Conversation.
In other words, for all dialogue tasks, the errors were smaller than the granularity of the evaluation scores (1.000).
This confirms the feasibility of the proposed evaluation framework and the validity of users' behaviors selected for this study.

\section{Conclusion} ~\label{sec:conclusion}

In this paper, we proposed an objective evaluation method for spoken dialogue systems used in social dialogue tasks.
We examined the relationship between users' behaviors and subjective evaluation scores for three different dialogue tasks: attentive listening, job interview, and first-meeting conversation.
Our findings revealed that the behaviors associated with the subjective evaluations vary depending on the characteristics of the dialogue task.
Additionally, we evaluated the proposed evaluation method framework through cross-validation and found that it can accurately predict subjective evaluation scores from the users' behaviors, with an absolute error smaller than the evaluation score’s granularity.
Moving forward, our future work aims to expand the analysis by considering more target behaviors and dialogue tasks.
We also plan to further develop the proposed evaluation method framework for other research and development purposes.

\section*{Acknowledgement}

This work was supported by JST ACT-X (JPMJAX2103), JST Moonshot R\&D (JPMJPS2011), and JSPS KAKENHI (JP19H05691 and JP23K16901).
The authors also express appreciation to Dr. Masato Komuro for his insightful comments on this research.

\bibliographystyle{unsrt}
\bibliography{IWSDS2024_KI_EVAL}

\end{document}